\documentclass[10pt,twocolumn,letterpaper]{article}

\usepackage{iccv}
\usepackage{times}
\usepackage{epsfig}
\usepackage{graphicx}
\usepackage{amsmath}
\usepackage{amssymb}

\usepackage{bm}
\usepackage{amssymb}
\usepackage{amsbsy}
\usepackage{multirow}
\usepackage{threeparttable}
\usepackage{dsfont}
\usepackage[switch]{lineno}
\usepackage[pagebackref=false,breaklinks=true,letterpaper=true,colorlinks,urlcolor=blue,citecolor=blue,linkcolor=blue,bookmarks=false]{hyperref}
\usepackage[accsupp]{axessibility}  

\iccvfinalcopy 

\ificcvfinal\fi

\begin{document}

\title{SignBERT: Pre-Training of Hand-Model-Aware Representation for Sign Language Recognition}

\author{Hezhen Hu{\small $~^{1}$}, ~Weichao Zhao{\small $~^{1}$}\thanks{Contribute equally with the first author.}, ~Wengang Zhou{\small $~^{1,2}$}\thanks{Corresponding author: Wengang Zhou and Houqiang Li.}, ~Yuechen Wang{\small $~^{1}$}, ~Houqiang Li{\small $~^{1,2\dag}$}\\
\normalsize
$^{1}$ CAS Key Laboratory of GIPAS, EEIS Department, University of Science and Technology of China (USTC) \\
\normalsize
$^{2}$ Institute of Artificial Intelligence, Hefei Comprehensive National Science Center\\

\normalsize
{\tt\small \{alexhu, saruka, wyc9725\}@mail.ustc.edu.cn, \{zhwg, lihq\}@ustc.edu.cn}
}

\maketitle
\ificcvfinal\thispagestyle{empty}\fi

\begin{abstract}
Hand gesture serves as a critical role in sign language.
Current deep-learning-based sign language recognition~(SLR) methods may suffer insufficient interpretability and overfitting due to limited sign data sources.
In this paper, we introduce the first self-supervised pre-trainable SignBERT with incorporated hand prior for SLR.
SignBERT views the hand pose as a visual token, which is derived from an off-the-shelf pose extractor.
The visual tokens are then embedded with gesture state, temporal and hand chirality information.
To take full advantage of available sign data sources, SignBERT first performs self-supervised pre-training by masking and reconstructing visual tokens.
Jointly with several mask modeling strategies, we attempt to incorporate hand prior in a model-aware method to better model hierarchical context over the hand sequence.
Then with the prediction head added, SignBERT is fine-tuned to perform the downstream SLR task.
To validate the effectiveness of our method on SLR, we perform extensive experiments on four public benchmark datasets, \emph{i.e.,} NMFs-CSL, SLR500, MSASL and WLASL.
Experiment results demonstrate the effectiveness of both self-supervised learning and imported hand prior.
Furthermore, we achieve state-of-the-art performance on all benchmarks with a notable gain.
\vspace{-0.3cm}
\end{abstract}

\section{Introduction}
Sign language, as a visual language, is the primary communication tool for the deaf community.
To facilitate the communication between the deaf and hearing people, sign language recognition~(SLR) has been widely studied with broad social influence.
Isolated SLR serves as a fundamental task in visual sign language research.
It aims to recognize sign language at the word-level and is a challenging fine-grained classification problem.

\begin{figure}
	\centering
	\includegraphics[width=1.0\linewidth]{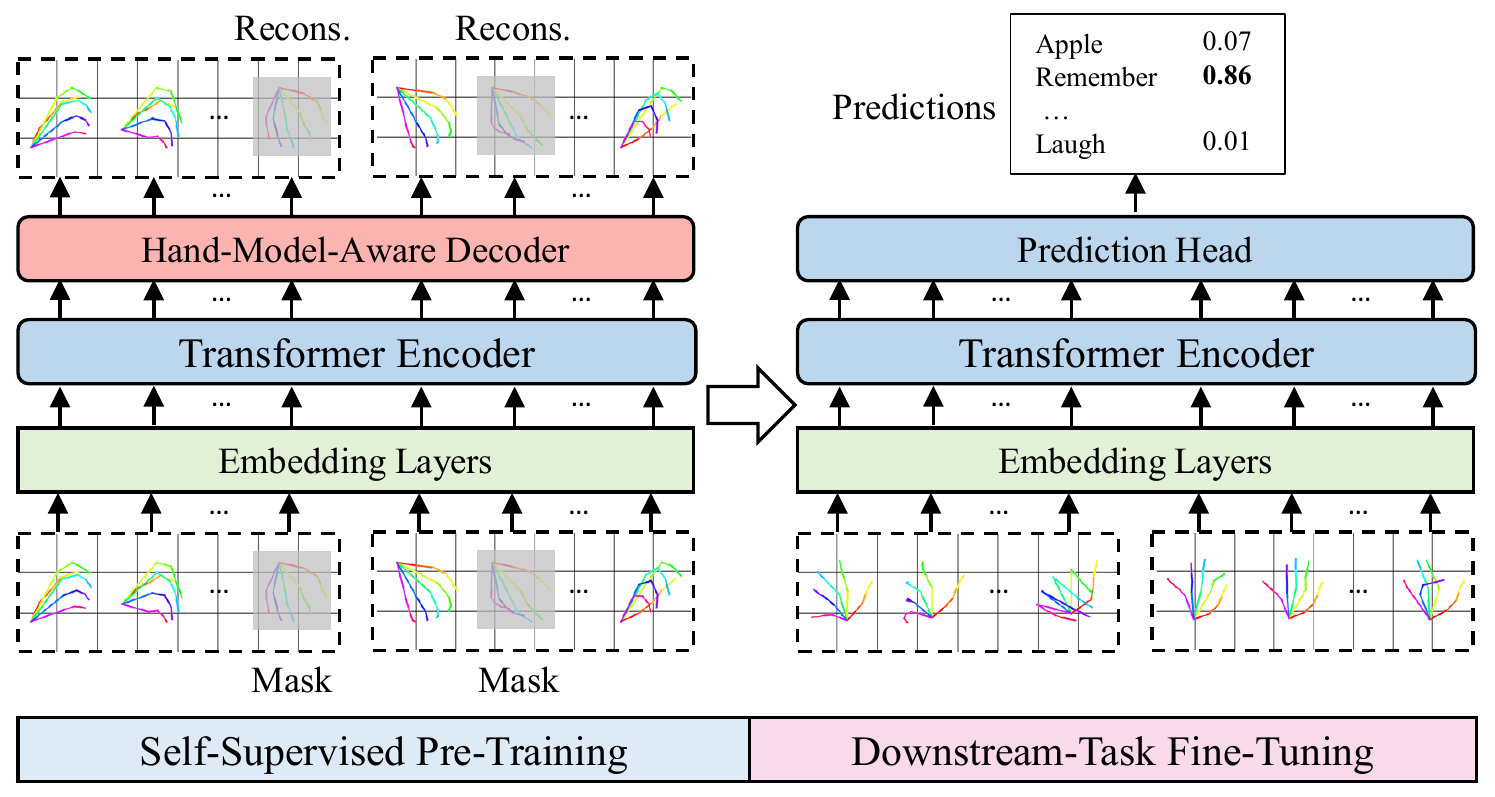}
	\caption{The overview of our framework, which contains self-supervised pre-training and downstream-task fine-tuning.}
	\label{fig:intro}
	\vspace{-0.4cm}
\end{figure}

Hand gesture serves as a dominant role during the expression of sign language.
It occupies a relatively small area with dynamic backgrounds, exhibits similar appearance and encounters self-occlusion among joints.
Such fact leads to the difficulty in hand representation learning.
Current deep-learning-based methods~\cite{camgoz2017subunets,koller2019weakly,huang2018video} learn feature representations adaptively from the cropped RGB hand sequence.
Given the highly articulated characteristic of hand, some methods represent them as sparse poses for recognition~\cite{albanie2020bsl,li2020word,joze2018ms}.
Pose is a compact and semantic representation, which is robust to appearance change and brings potential computation efficiency.
However, hand poses are usually extracted from the off-the-shelf extractor, which suffers failure detection.
Therefore, the performance of pose-based methods lags largely behind RGB-based counterparts.
Besides, the aforementioned methods all follow a data-driven paradigm and may suffer insufficient interpretability and overfitting due to limited sign data sources.

Meanwhile, the effectiveness of pre-training has been validated for computer vision~(CV) and natural language processing~(NLP).
Recent advance in NLP is largely derived from self-supervised pre-training strategies on large text corpus~\cite{radford2018improving,devlin2018bert,yang2019xlnet}.
Among them, BERT~\cite{devlin2018bert} is one of the most popular methods due to its simplicity and superior performance.
Its success is largely attributed to the powerful attention-based Transformer backbone~\cite{vaswani2017attention}, jointly with a well-designed pre-training strategy for modeling context inherent in text sequence.

To tackle the aforementioned issues, we develop a self-supervised pre-trainable framework with model-aware hand prior incorporated, namely SignBERT, as shown in Figure~\ref{fig:intro}.
Considering the compactness and expressiveness of hand pose representation, we view hand pose as a visual token.
Each hand token is embedded with gesture state, temporal and hand chirality information, and both hands are involved as input.
SignBERT first performs self-supervised pre-training on a large volume of hand pose data, which is derived from sign language data sources using the off-the-shelf extractor.
Specifically, inspired by BERT~\cite{devlin2018bert}, we pre-train our framework on the encoder-decoder backbone by masking and reconstructing visual tokens.
We design several mask modeling strategies to enforce the network capturing hierarchical contextual information.
To better capture context and ease optimization, the decoder introduces hand prior in a model-aware method.
For the downstream isolated SLR, the pre-trained encoder is fine-tuned with the added prediction head to perform recognition.

Our contributions are summarized as follows,
\vspace{-0.26cm}
\begin{itemize}
\setlength{\itemsep}{0pt}
\setlength{\parsep}{0pt}
\setlength{\parskip}{0pt}
    \item To our best knowledge, we propose the \emph{first} model-aware pre-trainable framework for sign language recognition, namely SignBERT. 
    It performs self-supervised learning on a large volume of hand pose data for better performance on the downstream task.
    \item To better exploit hierarchical contextual information contained in the sign data sources, we design mask modeling strategies and incorporate model-aware hand prior during self-supervised pre-training.
    \item We perform extensive experiments to validate the feasibility of our framework and its effectiveness on the downstream SLR task.
    Our method achieves state-of-the-art performance on four popular benchmarks, \emph{i.e.,} NMFs-CSL, SLR500, MSASL and WLASL.
\end{itemize}
\vspace{-0.2cm}

\section{Relate Work}
In this section, we will briefly review the related topics, including sign language recognition, pre-training strategy and hand-modeling technique.

\subsection{Sign Language Recognition}
Previous works~\cite{koller2020quantitative} on sign language recognition are generally grouped into two categories based on the input modality, \emph{i.e.}, RGB-based~(using the RGB video) and pose-based~(using the pose sequence) methods.

\noindent \textbf{RGB-based methods.} 
With the strong representation capability of CNNs, many works in SLR adopt it as the backbone~\cite{cheng2020fully,koller2018deep,joze2018ms,zhou2021improving}.
Necati~\emph{et al.}~\cite{camgoz2020sign} introduce a network consisting of 2D-CNNs for spatial representation and Transformer for modeling temporal dependencies by supervised learning. 
Some other works~\cite{huang2018attention, joze2018ms, li2020transfer, li2020word, albanie2020bsl} utilize 3D-CNNs for modeling spatio-temporal information.

\noindent \textbf{Pose-based methods.} 
As compact and semantic-aware data, pose sequences are processed by CNNs~\cite{li2018co, cao2018skeleton, albanie2020bsl} or RNNs~\cite{du2015hierarchical,min2020efficient,song2017end}.
Considering its well-structured nature, more and more works represent it as a graph and adopt graph convolutional networks~(GCNs) to model its representation~\cite{du2015hierarchical, song2017end,tunga2020pose}. 
Yan~\emph{et al.} \cite{yan2018spatial} first propose a spatial-temporal GCN for action recognition.
These GCN-based methods show both efficiency and promising performance.
There also exists work combining Transformer without pre-training for SLR~\cite{tunga2020pose}.

\subsection{Pre-Training Strategy}
Pre-training, a common strategy in NLP and CV, produces more generic feature representation and may alleviate overfitting for target tasks.
In NLP tasks, early works focused on improving word embedding~\cite{pennington2014glove,kiros2015skip}.
With the advance of Transformer~\cite{vaswani2017attention}, many works propose to pre-train generic feature representations~\cite{devlin2018bert, radford2018improving, yang2019xlnet}.
Of them, BERT is one of the most popular methods due to its simplicity and superior performance.
Specifically, two tasks are adopted in BERT pre-training, \emph{i.e.,} masked language modeling~(MLM) and next sentence prediction~(NSP).
In MLM, BERT attempts to predict the masked words based on the cues from unmasked context words.
In NSP, it defines a binary classification problem, which tries to predict whether two input sentences are consecutive.

In CV counterparts, it is common to pre-train the backbone on ImageNet~\cite{deng2009imagenet}, Kinetics~\cite{carreira2017quo} or large web sources~\cite{duan2020omni} for the downstream tasks.
There also exist works attempting to leverage the idea of BERT to CV tasks~\cite{sun2019videobert,Su2020VLBERT,li2020unicoder,Zhu_2020_CVPR,chen2020generative}.
In sign language, Albanie~\emph{et al.}~\cite{albanie2020bsl} propose to pre-train on a large annotated dataset and directly fine-tune on a small-scale one.
Li~\emph{et al.}~\cite{li2020transfer} fertilize recognition models by transferring knowledge of subtitled news sign videos to them.
To our best knowledge, there exists no work focusing on the self-supervised pre-training for SLR.

\subsection{Hand-Modeling Technique}
There have been many works to model the hand using various techniques, including sum-of-Gaussians~\cite{sridhar2013interactive}, shape primitives~\cite{oikonomidis2014evolutionary, qian2014realtime} and sphere-meshes~\cite{tkach2016sphere}. 
In order to model the hand shape more precisely, some works~\cite{ballan2012motion, tzionas2016capturing} propose to utilize a triangulated mesh with Linear Blend Skinning~(LBS)~\cite{lewis2000pose}. 
Recently, MANO~\cite{romero2017embodied} has become the most popular model with successful applications~\cite{habermann2020deepcap, boukhayma20193d, hu2021model, hu2021hand}. 
As a statistical model, MANO is learned from a large volume of high-quality hand scans.
Considering its capability of representing hand geometric changes in the low-dimensional shape and pose space, we adopt it as a constraint in the pose decoder to import hand prior.

\begin{figure*}
	\centering
	\includegraphics[width=1.0\linewidth]{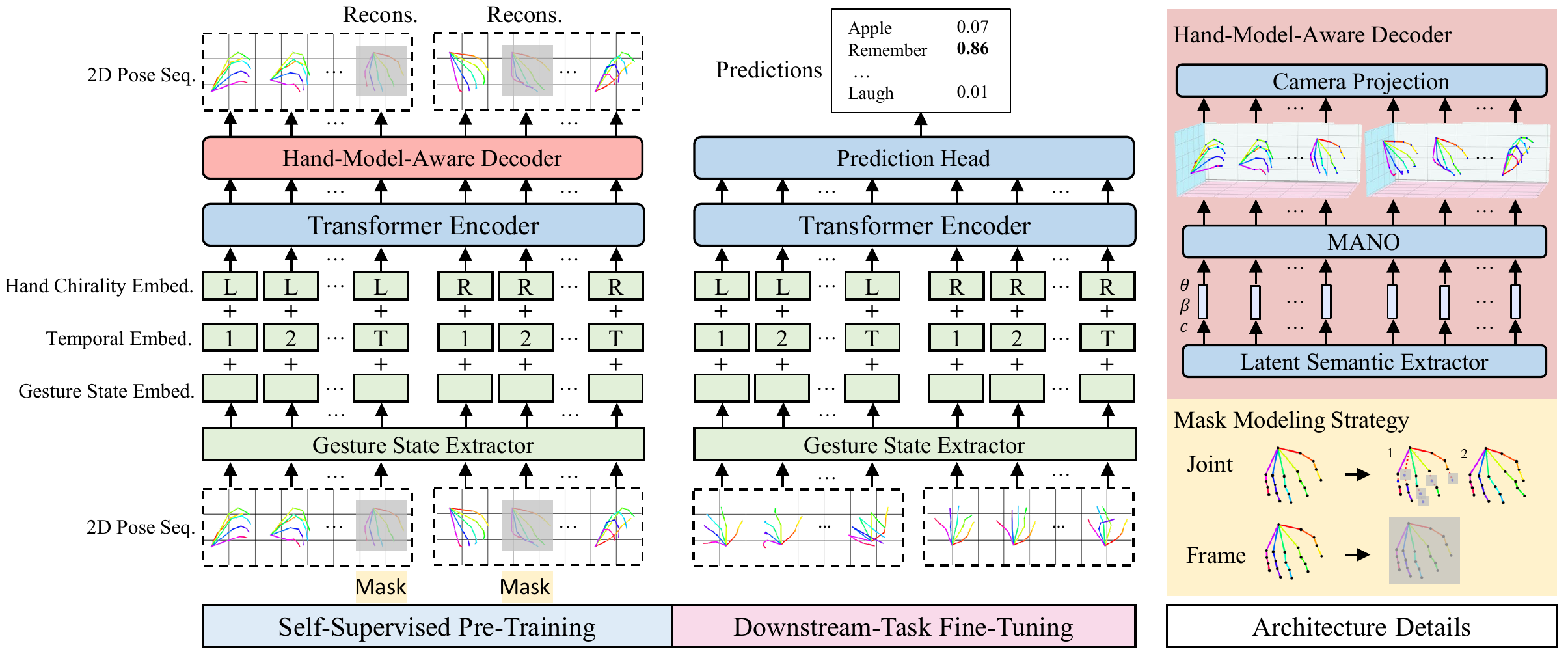}
	\caption{Illustration of our SignBERT framework, which contains self-supervised pre-training and fine-tuning for the downstream sign language recognition. The pre-extracted 2D hand pose sequence of both hands is fed into the framework. Each hand pose is viewed as a visual token, embedded with gesture state, temporal and hand chirality information. In self-supervised pre-training, we design several mask modeling strategies and incorporate model-aware hand prior to better exploit hierarchical contextual representation. For the downstream SLR task, the pre-trained Transformer encoder is fine-tuned with the prediction head to perform recognition.}
	\label{fig:overview}
	\vspace{-0.2cm}
\end{figure*}

\section{Our Approach}

\noindent \textbf{Overview.}
As shown in Figure~\ref{fig:overview}, SignBERT contains two stages, \emph{i.e.,} pre-training for modeling context in sign videos and fine-tuning for the downstream SLR task.
The hand poses, as visual tokens, are embedded with their gesture state, temporal and hand chirality information.
Since sign language is performed by two hands, we jointly feed them into our framework.
During pre-training, the whole framework works in a self-supervised paradigm by masking and reconstructing visual tokens.
Jointly with the mask modeling strategies, the decoder incorporates hand prior for better capturing hierarchical context of both hands and temporal dependencies during the sign.
When applying SignBERT to downstream recognition task, the hand-model-aware decoder is replaced by the prediction head, which is learned in a supervised paradigm by the corresponding video label.

In the following, we will first elaborate each component of our framework.
Then we will describe the proposed pre-training and fine-tuning procedures, respectively.

\subsection{Framework Architecture}
The hand pose in each frame is viewed as a visual token.
For each visual token, its input representation is constructed by summing the corresponding gesture state, temporal and hand chirality embeddings.

\noindent \textbf{Gesture state embedding $f_p$.}
Since the hand pose is well-structured with the physical connection among joints, we organize it as a spatial graph.
In this work, we adopt the spectral-based GCN from~\cite{cai2019exploiting, yan2018spatial} with a few modifications.
Given a 2D hand pose $\widetilde{J}_{t}$ representing the 2D location~(x and y coordinates) at frame $t$, an undirected spatial graph is defined by the node $V$ and edge $E$ set, respectively.
The node set includes all the corresponding hand joints, while the edge set contains the physical and symmetrical connections.
The hand pose sequence is first fed into several graph convolutional layers frame-by-frame.
Then graph pooling is performed based on neighbors to generate the frame-level semantics representation $f_{p,t}$.

\noindent \textbf{Temporal embedding $f_o$.}
Temporal information matters in video-level SLR.
Since self-attention does not consider the order information, we add the temporal order information by utilizing the position encoding strategy in~\cite{vaswani2017attention}.
Specifically, for the same hand, we add different temporal embeddings for different moments.
Meanwhile, since two hands simultaneously convey the meaning during sign, we add the same temporal embedding for the same moment, regardless of hand chirality.

\noindent \textbf{Hand chirality embedding $f_h$.}
Considering the meaning of sign language is conveyed by both hands, we introduce two special tokens to represent hand chirality of each frame, \emph{i.e.,} `L' and `R' for the left and right hand, respectively.
Specially, it is implemented by the WordPiece embeddings~\cite{wu2016google} with the same dimension as the gesture state and temporal embedding.
Notably, all the frames belonging to the same hand contain the identical hand chirality embedding.

\noindent \textbf{Transformer encoder.}
Given the aforementioned embedding representing the gesture status, temporal index and hand chirality, we sum them and feed it into the Transformer encoder following the original architecture~\cite{vaswani2017attention}, which contains a multi-head attention module and a feed forward network.
The encoder output $\mathbf{F}_N$, which retains the same size with the input, is computed as follows,
\begin{equation}
\begin{split}
	\mathbf{F}_0 &= \{f_{p}+f_{o}+f_{h}\}, \\
	\widetilde{\mathbf{F}}_i &= L(M(\mathbf{F}_{i-1}) + \mathbf{F}_{i-1}), \\
	\mathbf{F}_i &= L(C(\widetilde{\mathbf{F}}_i) + \widetilde{\mathbf{F}}_i),
\end{split}
\end{equation}
where $i$ denotes the $i$-th layer of the Transformer encoder, and we utilize totally $N$ layers. 
$L(\cdot)$, $M(\cdot)$ and $C(\cdot)$ denote the layer normalization, multi-head self-attention and feed forward network, respectively.
$\mathbf{F}_i$ denotes the feature representation in $i$-th layer.

\noindent \textbf{Hand-model-aware decoder.}
In our self-supervised pre-training paradigm, the framework needs to reconstruct the masked input sequence, in which the hand-model-aware decoder converts the feature to the pose sequence.
Specifically, a fully-connected layer $D(\cdot)$ first extracts a latent semantic embedding describing the hand status and camera parameters from the representation generated by the Transformer encoder, which is formulated as follows,
\begin{equation}
\label{equ:cnn_tcn}
  \mathbf{F}_{la} = \{\bm{\theta}, \bm{\beta}, \mathbf{c}_r, \mathbf{c}_o, c_s\}_{t=1}^T = D(\mathbf{F}_N),
\end{equation} 
where $\bm{\theta} \in \mathbb{R}^{25}$ and $\bm{\beta} \in \mathbb{R}^{10}$ are the pose and shape embedding for the following MANO, while $\mathbf{c}_r \in \mathbb{R}^{3\times3}$, $\mathbf{c}_o \in \mathbb{R}^{2}$, and $c_s \in \mathbb{R}$ are the weak-perspective camera parameters, indicating the rotation, translation and scale, respectively.

Then MANO~\cite{romero2017embodied} imports hand prior in a model-aware method and decodes the latent semantic embedding to hand representation.
MANO is a fully-differentiable model providing a mapping from low-dimensional pose $\bm{\theta}$ and shape $\bm{\beta}$ space to the triangulated hand mesh $\mathbf{M} \in \mathbb{R}^{N_v \times 3}$ with $N_v=$778 vertices and $N_f=$1538 faces.
To produce a physically plausible mesh, the pose and shape are constrained in a PCA space learned from a large volume of hand scan data.
The decoding process is formulated as follows,
\begin{equation}
\label{equ:mano}
  \mathbf{M}(\bm{\beta}, \bm{\theta}) = W(\mathbf{T}(\bm{\beta}, \bm{\theta}), J(\bm{\beta}), \bm{\theta}, \mathbf{W}),
  \vspace{-0.5cm}
\end{equation}
\begin{equation}
\label{equ:mano2}
  \mathbf{T}(\bm{\beta}, \bm{\theta}) = \bar{\mathbf{T}} + B_S(\bm{\beta}) + B_P(\bm{\theta}),
\end{equation}
where $\mathbf{W}$ is a set of blend weights. 
$B_S(\cdot)$ and $B_P(\cdot)$ denote shape and pose blend functions, respectively. 
The hand template $\bar{\mathbf{T}}$ is first posed and skinned based on the pose and shape corrective blend shapes, \emph{i.e.,} $B_P(\bm{\theta})$ and $B_S(\bm{\beta})$,
Then the mesh is generated by rotating each part around joints $J(\bm{\beta})$ using the linear skinning function $W(\cdot)$~\cite{kavan2005spherical}.
Besides, we are able to extract sparse 3D joints $\widetilde{J}_{3D}$ from the mesh.
To keep consistent with the widely-used hand annotation format, we further add 5 extra vertices with the index of 333, 443, 555, 678 and 734 as the fingertips, leading to total 21 3D joints.
Based on the predicted camera parameter, the predicted 3D joints are projected to the 2D plane.
The projected 2D hand pose is derived as follows,
\begin{equation}
\label{equ:weak}
  \widetilde{J}_{2D} = c_s\prod{({\mathbf{c}_{r}}{{\widetilde{J}}_{3D}})}+\mathbf{c}_o,
\end{equation} 
where $\prod(\cdot)$ denotes the orthographic projection.

\noindent \textbf{Prediction head.}
Since discriminative cues may only contain in certain frames, we utilize a simple attention mechanism to weight features temporally.
Then the weighted features are summed to perform final classification.

\subsection{Pre-Training SignBERT}
In this section, we elaborate SignBERT pre-training paradigm on a large volume of sign data sources to exploit semantic context hierarchically.
Different from the original BERT pre-training on discrete word space, we aim to pre-train on continuous hand pose space.
Substantially, the classification problem is transformed into regression, which poses new challenges on the reconstruction of the hand pose sequence.
To tackle this issue, we view hand poses as visual `words' (continuous tokens) and jointly utilize the aforementioned model-aware decoder as a constraint with hand prior incorporated.
Given a hand sequence containing both hands, we first randomly choose 50\% tokens.
Similar to BERT, if the token is chosen, we randomly perform one of three operations with equal probability, \emph{i.e.,} masked joint modeling, masked frame modeling and identity modeling.

\noindent \textbf{Masked joint modeling.}
Since current pose detectors may contain failure detection on some joints, we incorporate masked joint modeling to mimic the usual failure cases.
In a chosen token, we randomly choose $m$ joints ranging from 1 to $M$.
For these chosen joints, we perform two operations with equal probability, \emph{i.e.,} zero masking~(masking the coordinates of joints with zeros) or random spatial disturbance.
This modeling attempts to embed our framework the capability to infer the gesture state from remaining hand joints, thus capturing context at the joint level.

\noindent \textbf{Masked frame modeling.}
Masked frame modeling is performed on a more holistic view.
For a chosen token, all the joints are zero masked.
The framework is enforced to reconstruct this token by observations from remaining pose tokens of the other hand or different temporal points.
In this way, temporal context in each hand and mutual context between hands are captured.

\noindent \textbf{Identity modeling.}
Identity modeling makes the unchanged token fed into the framework.
This operation is indispensable for the framework to learn identity mapping on those unmasked tokens.

\subsection{Objective Functions in Pre-Training}
The proposed three strategies allow the network to maximize the likelihood of the joint probability distribution to reconstruct the hand pose sequence.
In this manner, the context contained in the sequence is captured.
During pre-training, only the output corresponding to chosen tokens are included in the following loss calculation as follows,
\begin{equation}
\label{equ:pre-train}
  \mathcal{L} = \mathcal{L}_{rec} + \lambda \mathcal{L}_{reg},
\end{equation}
where $\lambda$ denotes the weighting factor.

\noindent \textbf{Hand reconstruction loss $\mathcal{L}_{rec}$.}
Since hand pose detection results $J_{2D}$ serve as the pseudo label, we ignore the joints with the prediction confidence lower than $\epsilon$ and utilize the remaining joints weighted by the confidence in the calculation of this loss term.
\begin{equation}
\small
\label{equ:rec}
  \mathcal{L}_{rec} = \sum\limits_{t, j}\mathds{1}(c(t,j)>=\epsilon)c(t,j){{\left\|\widetilde{J}_{2D}(t,j) - J_{2D}(t,j) \right\|}_{1}},
\end{equation}
where $\mathds{1}(\cdot)$ denotes the indicator function, and $c(t,j)$ denotes the confidence of the ${J}_{2D}$ with joint $j$ at time $t$.

\noindent \textbf{Regularization loss $\mathcal{L}_{reg}$.}
To ensure the hand model working properly, a regularization loss is added.
It is implemented by constraining magnitude and derivative of the MANO input, which is responsible for generating the plausible mesh and keeping the signer identity unchanged.
The regularization loss is calculated as follows,
\begin{equation}
\label{equ:reg}
  \mathcal{L}_{reg} = \sum\limits_{t}( {\left\|\theta_t\right\|}_{2}^2 + w_{\beta}{\left\|\beta_t \right\|}_{2}^2 + w_{\delta}{\left\|\beta_{t} - \beta_{t-1} \right\|}_{2}^{2}),
\end{equation}
where $w_{\beta}$ and $w_{\delta}$ denote the weighting factor.

\subsection{Fine-Tuning SignBERT}
After pre-training SignBERT, it is relatively simple to fine-tune it for the downstream SLR task.
The hand-model-aware decoder is replaced by the prediction head.
The input hand pose sequence is all unmasked and we use the cross-entropy loss to supervise the output of the prediction head.

Considering only the hand pose sequence is insufficient to convey the full meaning of sign language, it is necessary to fuse recognition results based on hands with that of full frame.
The full frame can be represented by full RGB data or full keypoints. 
In our work, we use the simple late fusion strategy, which directly sums their prediction results.
Besides, the full RGB and keypoints baseline method utilized for fusion are marked in each dataset for clarity.
In the following, we refer our method with only hands, fusion of hands and full RGB data, fusion of hands and full keypoints as \textbf{Ours~(H)}, \textbf{Ours~(H + R)} and \textbf{Ours~(H + P)}, respectively.

\section{Experiments}
\subsection{Datasets and Evaluation}
\noindent
\textbf{Datasets.} 
We evaluate our proposed method on four public sign language datasets, including NMFs-CSL \cite{hu2020global}, SLR500 \cite{huang2018attention}, MSASL \cite{joze2018ms} and WLASL \cite{li2020word}. 

\textbf{NMFs-CSL} is the most challenging Chinese sign language~(CSL) dataset due to a large variety of confusing words caused by fine-grained cues. 
It totally contains 1,067 words with 610 confusing words and 457 normal words. 
There are 25,608 and 6,402 samples for training and testing, respectively.
\textbf{SLR500} is another CSL dataset, which contains 500 daily words with 125,000 recording samples performed by 50 signers. 
Specifically, 90,000 and 35,000 samples are utilized for training and testing, respectively.

\textbf{MSASL} is an American sign language dataset~(ASL) containing a vocabulary size of 1,000, with 25,513 samples in total for training, validation and testing, respectively. 
Besides, the Top-100 and Top-200 most frequent words are chosen as its two subsets, referred to as MSASL100, MSASL200.
\textbf{WLASL} is another ASL dataset with a vocabulary of 2,000 words and 21,083 samples.
Similar to MSASL, it releases WLASL100 and WLASL300 as its subsets.
MSASL and WLASL are both collected from Web videos and bring new challenges due to unconstrained real-life recording conditions and limited samples for each word.

Meanwhile, since STB~\cite{zhang2017hand} and HANDS17~\cite{yuan20172017} provide 2D hand joint annotations, we utilize them to validate the feasibility of our proposed framework.

\textbf{STB} is a real-world hand pose estimation datasets, which contains 18,000 samples. 
Following Zimmermann~\emph{et al.} \cite{zimmermann2017learning}, we split this dataset into 15,000 training and 3,000 testing samples for single-frame validation.
\textbf{HANDS17} is a video-level hand pose estimation dataset, containing a total of 292,820 frames from 99 video sequences. 
In this dataset, we split the first 70$\%$ and last 30$\%$ frames in each sequence for training and testing, respectively.

\noindent
\textbf{Evaluation.} 
For the downstream isolated SLR task, we utilize the accuracy metrics, \emph{i.e.,} the per-class~(\textbf{P-C}) and per-instance~(\textbf{P-I}) metrics, which denote the average accuracy over each class and each instance, respectively. 
We report the Top-1 and Top-5 accuracy under both per-instance and per-class for MSASL and WLASL.
Since NMFs-CSL and SLR500 contain the same number of samples for each class, we only report per-instance accuracy following~\cite{hu2020global, huang2018attention}.

For STB and HANDS17, we report the Percentage of Correct Keypoints (PCK) score and the area under the curve (AUC) on the PCK ranging from 20 to 40 pixels, which are widely-used criteria to evaluate pose estimation accuracy. 
Specifically, PCK defines a candidate keypoint to be correct if it falls within a circle (2D) of a given radius around the ground truth, where the distances are expressed in pixels.

\subsection{Implementation Details}
In our experiment, all the models are implemented by PyTorch~\cite{paszke2019pytorch} and trained on NVIDIA RTX 3090. 
Since no pose annotation is available in sign language datasets, we use MMPose~\cite{mmpose2020} for its efficiency to extract the 133 full 2D keypoints, \textit{i.e.}, the 23 body joints, 68 face and 42 hand joints.
The extracted hand and shoulder joints are further utilized to crop the left and right hand pose and rescale them to 256 $\times$ 256.
Both hands are fed into the framework.
The framework is trained with the Adam optimizer. 
The weight decay and momentum are set to 0.0001 and 0.9, respectively.
We start at the initial learning rate of 0.001 and reduce it by a factor of 0.1 every 20 epochs.
In all experiments, the hyper parameters $\epsilon$, $\lambda$, $w_{\beta}$ and $w_{\delta}$ are set as 0.5, 0.01, 10.0 and 100.0, respectively.
During the pre-training stage, we include the training data from all four aforementioned sign language datasets.
For the downstream task, we temporally extract 32 frames using random and center sampling during training and testing, respectively.

\subsection{Ablation Study}
In this section, we first validate the feasibility of our framework.
Then we perform ablation studies to demonstrate the effectiveness of the main components in our framework.

\noindent \textbf{Framework feasibility.}
We validate the feasibility of our framework on the datasets with hand pose annotation available.
As shown in Table~\ref{STB}, we first validate reconstruction ability under the single-frame setting on the STB dataset.
Specifically, a single frame is fed into the framework.
We only perform the masked joint modeling, where $M$ indicates the number of masked joints ranges from 1 to $M$, resulting the average number as $M/2$.
With the gradual increase of $M$, the PCK and AUC metrics of reconstructed joints are consistently higher than those of the input.
It demonstrates that our framework is able to hallucinate the whole hand pose by observing partial joints.

\begin{table}
\small
\tabcolsep=11pt
\begin{center}
\begin{tabular}{c|cc|cc}
\hline
\multirow{2}{*}{M}  & \multicolumn{2}{c|}{Input} & \multicolumn{2}{c}{Output} \\
   & P@20  & AUC   & P@20  & AUC \\ \hline \hline
3  & 88.81 & 91.02 & 99.90 & 99.54  \\
5  & 82.26 & 85.65 & 99.89 & 99.53  \\
7  & 76.19 & 80.91 & 99.85 & 99.53  \\
9  & 70.85 & 76.63 & 99.81 & 99.50  \\
11 & 66.29 & 72.85 & 99.79 & 99.44  \\ \hline
\end{tabular}
\end{center}
\caption{Frame-level framework feasibility on the STB dataset. `P@20' denotes the PCK metrics with the error threshold set as 20 pixel. We only utilize the masked joint modeling, and $M$ denotes the number of masked joints ranges from 1 to $M$.}
\label{STB}
\vspace{-0.3cm}
\end{table}

\begin{table}
\small
\tabcolsep=6.5pt
\begin{center}
\begin{tabular}{cc|cc|cc}
\hline
\multicolumn{2}{c|}{Mask} & \multicolumn{2}{c|}{Input} & \multicolumn{2}{c}{Output} \\
Joint       & Frame        & P@20  & AUC   & P@20  & AUC   \\ \hline \hline
\checkmark  &              & 86.38 & 89.02 & 95.13 & 95.49 \\
            & \checkmark   & 80.85 & 80.85 & 95.33 & 95.57 \\
\checkmark  & \checkmark   & 81.43 & 82.32 & 95.14 & 95.48 \\ \hline          
\end{tabular}
\end{center}
\caption{Video-level framework feasibility on HANDS17. `P@20' denotes the PCK metrics with the error threshold set as 20 pixel. `Joint' and `Frame' denote the masked joint modeling and masked frame modeling, respectively.}
\label{HANDS17}
\vspace{-0.2cm}
\end{table}

\begin{figure}
	\centering
	\includegraphics[width=1.0\linewidth]{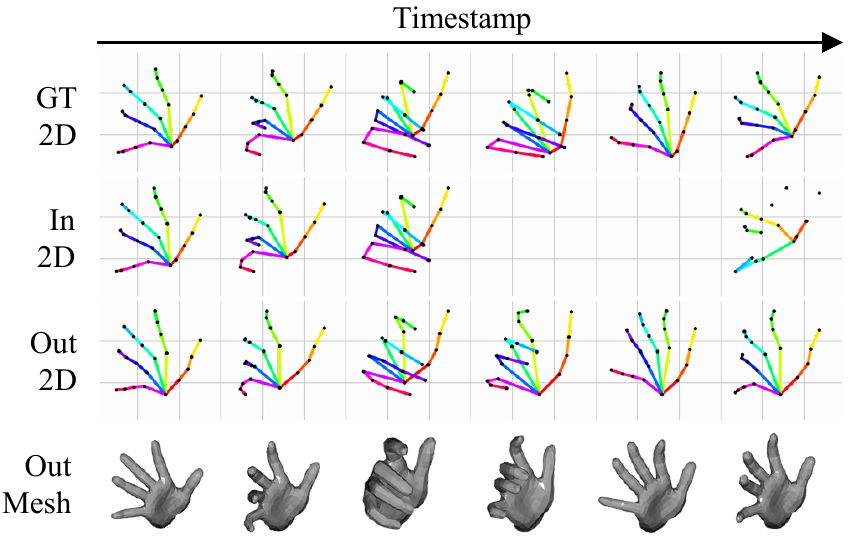}
	\caption{Visualization of the framework feasibility on HANDS17. We choose 6 continuous frames from one video. The four rows denote the ground truth~(GT) pose sequence, input sequence after performing masking on GT, the reconstructed sequence and middle results of the mesh sequence, respectively. Notably, two blanks in the second row represent these poses are all masked. 
	}
	\vspace{-0.3cm}
	\label{fig:HANDS17}
\end{figure}

From Table~\ref{HANDS17}, the framework feasibility under the video-level setting is tested on the HANDS17 dataset. 
We utilize all masking strategies on the original pose sequence to formulate the input.
It can be also observed that the PCK and AUC performance of the output sequence are higher than those of input, which verifies the framework capability of reconstructing from inaccurate hand joint sequence.
Besides, we visualize the hand pose reconstruction in Figure~\ref{fig:HANDS17}.

\begin{table*}
\small
\tabcolsep=11.2pt
\begin{center}
\begin{threeparttable}
\begin{tabular}{l|ccc|ccc|ccc}
\hline
\multirow{2}{*}{Method}      &  \multicolumn{3}{c|}{Total} &  \multicolumn{3}{c|}{Confusing} &  \multicolumn{3}{c}{Normal}  \\ 
         & Top-1 & Top-2 & Top-5 & Top-1 & Top-2 & Top-5 & Top-1 & Top-2 & Top-5 \\ \hline \hline 
\textbf{Pose-based}  & & & & & & & & & \\
ST-GCN~\cite{yan2018spatial} & 59.9 & 74.7 & 86.8 & 42.2 & 62.3 & 79.4 & 83.4 & 91.3 & 96.7 \\
Ours~(H)  & 67.0 & 86.8 & 95.3 & 46.4 & 78.2 & 92.1 & 94.5 & 98.1 & 99.6 \\
Ours~(H + P) & \textbf{74.9} & \textbf{93.2} & \textbf{98.2} & \textbf{58.6} & \textbf{88.6} & \textbf{96.9} & \textbf{96.7} & \textbf{99.3} & \textbf{99.9} \\ \hline
\textbf{RGB-based}  & & & & & & & & & \\
3D-R50~\cite{qiu2017learning}  & 62.1 & 73.2 & 82.9 & 43.1 & 57.9 & 72.4 & 87.4 & 93.4 & 97.0 \\
DNF~\cite{cui2019deep}         & 55.8 & 69.5 & 82.4 & 33.1 & 51.9 & 71.4 & 86.3 & 93.1 & 97.0 \\
I3D~\cite{carreira2017quo}     & 64.4 & 77.9 & 88.0 & 47.3 & 65.7 & 81.8 & 87.1 & 94.3 & 97.3 \\
TSM~\cite{lin2019tsm}          & 64.5 & 79.5 & 88.7 & 42.9 & 66.0 & 81.0 & 93.3 & 97.5 & 99.0 \\
Slowfast~\cite{feichtenhofer2019slowfast} & 66.3 & 77.8 & 86.6 & 47.0 & 63.7 & 77.4 & 92.0 & 96.7 & 98.9 \\ 
GLE-Net~\cite{hu2020global}  & 69.0 & 79.9  & 88.1 & 50.6 & 66.7 & 79.6 & 93.6 & 97.6 & 99.3 \\
Ours~(H + R)  & \textbf{78.4} & \textbf{92.0} & \textbf{97.3} & \textbf{64.3} & \textbf{86.5} & \textbf{95.4} & \textbf{97.4} & \textbf{99.3} & \textbf{99.9} \\ \hline
\end{tabular}
\end{threeparttable}
\end{center}
\caption{Accuracy comparison on NMFs-CSL dataset. \cite{yan2018spatial} and \cite{qiu2017learning} denote the pose and RGB baseline, respectively.}
\label{NMFs-CSL}
\vspace{-0.3cm}
\end{table*}

\begin{table}
\small
\tabcolsep=2.8pt
\begin{center}
\begin{tabular}{cc|cc|cc|cc}
\hline
\multicolumn{2}{c|}{Mask} & \multicolumn{2}{c|}{100} & \multicolumn{2}{c|}{200} & \multicolumn{2}{c}{1000} \\
Joint      & Frame      & P-I & P-C & P-I & P-C & P-I & P-C \\ \hline \hline
           &            & 63.01 & 62.72  & 57.69 & 57.56 & 41.85 & 38.30  \\
\checkmark &            & 72.66 & 72.75  & 68.51 & 69.72 & 48.87 & 45.39 \\
           & \checkmark & 74.77 & 75.48  & 68.65 & 69.20 & 49.02 & 46.02 \\
\checkmark & \checkmark & \textbf{76.09} & \textbf{76.65} & \textbf{70.64} & \textbf{70.92} & \textbf{49.54} & \textbf{46.39} \\ \hline          
\end{tabular}
\end{center}
\caption{Effectiveness of the masking strategy on MSASL dataset. The first row denotes the baseline, \emph{i.e.,} our framework is trained without pre-training. `Joint' and `Frame' denote the masked joint modeling and masked frame modeling, respectively.}
\label{mask}
\vspace{-0.3cm}
\end{table}

Since we focus on the performance of the downstream recognition task, we perform extensive experiments on MSASL and its subsets to demonstrate the effectiveness of the masking strategies, model-aware decoder, Transformer layers $N$ and pre-training data scale.
We report per-instance and per-class Top-1 accuracy as the performance indicator.

\noindent \textbf{Effectiveness of the masking strategy.}
As illustrated in Table~\ref{mask}, the first row denotes the baseline method, \emph{i.e.,} our framework is directly trained under the video label supervision without pre-training.
It is worth mentioning that compared with this baseline, our designed pre-training brings notable performance gain, with 13.08\%, 12.95\% and 7.69\% Top-1 per-instance accuracy improvement.
Both joint-level and frame-level masking strategies are beneficial for the framework capturing different levels of context, thus bringing performance improvement.
When two masking strategies are both utilized, it reaches the best performance.

\begin{table}[t]
\small
\tabcolsep=4pt
\begin{center}
\begin{tabular}{c|cc|cc|cc}
\hline
\multicolumn{1}{c|}{\multirow{2}{*}{Decoder}} & \multicolumn{2}{c|}{100} & \multicolumn{2}{c|}{200} & \multicolumn{2}{c}{1000} \\
           & P-I & P-C & P-I   & P-C   & P-I   & P-C   \\ \hline \hline
1-layer fc & 73.05 & 72.62 & 67.55 & 68.21 & 47.94 & 45.07  \\
2-layer fc & 74.24 & 74.21 & 68.29 & 69.12 & 48.03 & 45.25   \\
Ours       & \textbf{76.09} & \textbf{76.65} & \textbf{70.64} & \textbf{70.92} & \textbf{49.54} & \textbf{46.39} \\ \hline   
\end{tabular}
\end{center}
\caption{Effectiveness of the model-aware decoder on MSASL dataset. We compare ours with different pose decoders.}
\label{model-aware}
\vspace{-0.2cm}
\end{table}

\noindent \textbf{Effectiveness of the model-aware decoder.}
As shown in Table~\ref{model-aware}, we compare the effect of different pose decoders on SLR.
The first two rows denote utilizing the fully-connected layers to regress the hand pose.
Our decoder work in a model-aware method to import hand prior during pre-training, which eases optimization and brings performance improvement for downstream isolated SLR.
Besides, the model-aware decoder has additional benefits, which inflates the 2D hand pose sequence to the 3D plane.

\begin{table}[t]
\small
\tabcolsep=6pt
\begin{center}
\begin{tabular}{c|cc|cc|cc}
\hline
\multicolumn{1}{c|}{\multirow{2}{*}{$N$}} & \multicolumn{2}{c|}{100} & \multicolumn{2}{c|}{200} & \multicolumn{2}{c}{1000} \\
      & P-I & P-C & P-I & P-C & P-I & P-C   \\ \hline \hline
2 & 74.11 & 74.61  & 67.70 & 67.92 & 48.23 & 45.17   \\
3 & \textbf{76.09} & \textbf{76.65} & \textbf{70.64} & \textbf{70.92} & \textbf{49.54} & \textbf{46.39} \\  
4 & 75.69 & 75.51 & 70.20 & 70.66 & 47.36 & 44.04  \\
5 & 74.90 & 75.68 & 68.14 & 68.40 & 47.29 & 44.42  \\ \hline  
\end{tabular}
\end{center}
\caption{Effectiveness of the Transformer layers $N$ on MSASL dataset. $N$ denotes the number of the layers in the Transformer encoder.}
\label{layer}
\vspace{-0.3cm}
\end{table}

\noindent \textbf{Effectiveness of Transformer layers $N$.}
From Table~\ref{layer}, the accuracy increases, when the number of Transformer layers increases.
It reaches the peak when $N=3$.
The difference of the best layers in BERT and our model may be due to different characteristics between sign pose and NLP domain, and the overfitting issue.
Unless stated, we utilize $N=3$ in all our experiments.

\noindent \textbf{Effectiveness of the pre-training data scale.}
As shown in Table~\ref{data-scale}, as the ratio of pre-training data volume increases, the performance on the downstream SLR task gradually increases on the accuracy metrics.
It indicated that SignBERT may benefit from larger pre-training datasets.

\begin{table}[t]
\small
\tabcolsep=5pt
\begin{center}
\begin{tabular}{c|cc|cc|cc}
\hline
\multicolumn{1}{c|}{\multirow{2}{*}{Ratio}} & \multicolumn{2}{c|}{100} & \multicolumn{2}{c|}{200} & \multicolumn{2}{c}{1000} \\
           & P-I & P-C & P-I   & P-C   & P-I   & P-C   \\ \hline \hline
0\%  & 63.01 & 62.72 & 57.69 & 57.56 & 41.85 & 38.30  \\
25\% & 73.18 & 72.83 & 67.91 & 69.30 & 46.18 & 43.97  \\
50\% & 73.18 & 73.42 & 67.18 & 67.71 & 46.57 & 43.79  \\
75\% & 74.50 & 74.36 & 68.72 & 68.97 & 47.21 & 43.67  \\
100\% & \textbf{76.09} & \textbf{76.65} & \textbf{70.64} & \textbf{70.92} & \textbf{49.54} & \textbf{46.39} \\ \hline   
\end{tabular}
\end{center}
\caption{Effectiveness of the ratio of pre-training data scale on the MSASL dataset.}
\label{data-scale}
\vspace{-0.2cm}
\end{table}

\begin{table*}
\small
\tabcolsep=7pt
\begin{center}
\begin{threeparttable}
\begin{tabular}{l|cc|cc|cc|cc|cc|cc}
\hline
\multirow{3}{*}{Method} & \multicolumn{4}{c|}{MSASL100}
                        & \multicolumn{4}{c|}{MSASL200}
                        & \multicolumn{4}{c}{MSASL1000}\\ \cline{2-13}
        & \multicolumn{2}{c|}{Per-instance} & \multicolumn{2}{c|}{Per-class} 
        & \multicolumn{2}{c|}{Per-instance} & \multicolumn{2}{c|}{Per-class}
        & \multicolumn{2}{c|}{Per-instance} & \multicolumn{2}{c}{Per-class} \\ 
        & Top-1 & Top-5 & Top-1 & Top-5 
        & Top-1 & Top-5 & Top-1 & Top-5 
        & Top-1 & Top-5 & Top-1 & Top-5    \\ \hline \hline
\textbf{Pose-based} & & & & & & & & & & & & \\
ST-GCN~\cite{yan2018spatial}  & 59.84 & 82.03 & 60.79 & 82.96  
        & 52.91 & 76.67 & 54.20 & 77.62
        & 36.03 & 59.92 & 32.32 & 57.15 \\
Ours~(H) & 76.09 & 92.87 & 76.65 & 93.06  
         & 70.64 & 89.55 & 70.92 & 90.00
         & 49.54 & 74.11 & 46.39 & 72.65 \\
Ours~(H + P) & \textbf{81.37} & \textbf{93.66} & \textbf{82.31} & \textbf{93.76} 
             & \textbf{77.34} & \textbf{91.10} & \textbf{78.02} & \textbf{91.48}
             & \textbf{59.80} & \textbf{81.86} & \textbf{57.06} & \textbf{80.94} \\ \hline
\textbf{RGB-based} & & & & & & & & & & & &   \\
I3D~\cite{joze2018ms}  & - & - & 81.76 & 95.16  
        & - & - & 81.97 & 93.79
        & - & - & 57.69 & 81.05 \\ 
TCK~\cite{li2020transfer}  & 83.04 & 93.46 & 83.91 & 93.52  
        & 80.31 & 91.82 & 81.14 & 92.24
        & - & - & - & - \\ 
BSL~\cite{albanie2020bsl}  & - & - & - & -  
        & - & - & - & -
        & 64.71 & 85.59 & 61.55 & 84.43  \\
Ours~(H + R)   & \textbf{89.56} & \textbf{97.36} & \textbf{89.96} & \textbf{97.51}
             & \textbf{86.98} & \textbf{96.39} & \textbf{87.62} & \textbf{96.43}
             & \textbf{71.24} & \textbf{89.12} & \textbf{67.96} & \textbf{88.40} \\ \hline 
\end{tabular}
\end{threeparttable}
\end{center}
\caption{Accuracy comparison on MSASL dataset. \cite{yan2018spatial} and \cite{joze2018ms} denote the pose and RGB baseline, respectively.}
\label{msasl}
\vspace{-0.2cm}
\end{table*}

\begin{table*}
\small
\tabcolsep=6.5pt
\begin{center}
\begin{threeparttable}
\begin{tabular}{l|cc|cc|cc|cc|cc|cc}
\hline
\multirow{3}{*}{Method} & \multicolumn{4}{c|}{WLASL100}
                        & \multicolumn{4}{c|}{WLASL300}
                        & \multicolumn{4}{c}{WLASL2000}\\ \cline{2-13}
        & \multicolumn{2}{c|}{Per-instance} & \multicolumn{2}{c|}{Per-class} 
        & \multicolumn{2}{c|}{Per-instance} & \multicolumn{2}{c|}{Per-class}
        & \multicolumn{2}{c|}{Per-instance} & \multicolumn{2}{c}{Per-class} \\
        & Top-1 & Top-5 & Top-1 & Top-5 
        & Top-1 & Top-5 & Top-1 & Top-5   
        & Top-1 & Top-5 & Top-1 & Top-5    \\ \hline \hline
\textbf{Pose-based} & & & & & & & & & & & & \\
ST-GCN~\cite{yan2018spatial} & 50.78 & 79.07 & 51.62 & 79.47   
        & 44.46 & 73.05 & 45.29 & 73.16
        & 34.40 & 66.57 & 32.53 & 65.45 \\ 
Pose-TGCN~\cite{li2020word} & 55.43 & 78.68 & - & -   
        & 38.32 & 67.51 & - & -
        & 23.65 & 51.75 & - & - \\ 
PSLR~\cite{tunga2020pose} & 60.15 & 83.98 & - & -   
        & 42.18 & 71.71 & - & -
        & - & - & - & - \\ 
Ours~(H) & 76.36 & 91.09 & 77.68 & 91.67  
         & 62.72 & 85.18 & 63.43 & 85.71 
         & 39.40 & 73.35 & 36.74 & 72.38 \\
Ours~(H + P) & \textbf{79.07} & \textbf{93.80} & \textbf{80.05} & \textbf{94.17} 
             & \textbf{70.36} & \textbf{88.92} & \textbf{71.17} & \textbf{89.36}
             & \textbf{47.46} & \textbf{83.32} & \textbf{45.17} & \textbf{82.32} \\ \hline
\textbf{RGB-based} & & & & & & & & & & & & \\
I3D~\cite{li2020word}  & 65.89 & 84.11 & 67.01 & 84.58  
        & 56.14 & 79.94 & 56.24 & 78.38
        & 32.48 & 57.31 & - & - \\ 
TCK~\cite{li2020transfer} & 77.52 & 91.08 & 77.55 & 91.42   
        & 68.56 & 89.52 & 68.75 & 89.41
        & - & - & - & - \\ 
BSL~\cite{albanie2020bsl}  & - & - & - & -  
        & - & - & - & -
        & 46.82 & 79.36 & 44.72 & 78.47 \\ 
Ours~(H + R)   & \textbf{82.56} & \textbf{94.96} & \textbf{83.30} & \textbf{95.00}
             & \textbf{74.40} & \textbf{91.32} & \textbf{75.27} & \textbf{91.72}
             & \textbf{54.69} & \textbf{87.49} & \textbf{52.08} & \textbf{86.93} \\ \hline 
\end{tabular}
\end{threeparttable}
\end{center}
\caption{Accuracy comparison on WLASL dataset. ST-GCN~\cite{yan2018spatial} and I3D~\cite{li2020word} denote the pose and RGB baseline, respectively.}
\label{wlasl}
\vspace{-0.3cm}
\end{table*}

\begin{table}
\small
\begin{center}
\tabcolsep=14pt
\begin{threeparttable}
\begin{tabular}{l|c}
\hline
Method  &  Accuracy   \\  \hline \hline 
\textbf{Pose-based} \\
ST-GCN~\cite{yan2018spatial} &  90.0 \\
Ours~(H)      & 94.5     \\ 
Ours~(H + P)  & \textbf{96.6}      \\ \hline
\textbf{RGB-based}  \\
STIP~\cite{laptev2005space}   &  61.8 \\
GMM-HMM~\cite{tang2015real} &  56.3 \\
3D-R50~\cite{qiu2017learning}  &  95.1 \\
GLE-Net~\cite{hu2020global}   & 96.8      \\  \hline
Ours~(H + R)  & \textbf{97.6}      \\ \hline 
\end{tabular}
\end{threeparttable}
\end{center}
\vspace{-0.2cm}
\caption{Accuracy comparison on SLR500 dataset. \cite{yan2018spatial} and \cite{qiu2017learning} denote the pose and RGB baseline, respectively. }
\label{slr500}
\vspace{-0.3cm}
\end{table}

\subsection{Comparison with State-of-the-art Methods}
We compare our method with previous state-of-the-art methods on four benchmark datasets.
For clarity, previous methods are grouped by their input modality, \emph{i.e.,} pose-based and RGB-based methods.

\noindent \textbf{Evaluation on NMFs-CSL.}
As illustrated in Table~\ref{NMFs-CSL}, we compare with methods~\cite{yan2018spatial, qiu2017learning, cui2019deep, carreira2017quo, lin2019tsm, feichtenhofer2019slowfast, hu2020global} utilizing the pose and RGB sequence as input. 
GLE-Net \cite{hu2020global} is the most challenging method, which enhances discriminative cues from global and local views. 
It is worth noting that our method with purely using hand pose achieves comparable performance with a majority of them. 
Ours~(H + R) outperforms all previous methods with a notable margin.

\noindent \textbf{Evaluation on SLR500.}
As shown in Table~\ref{slr500}, STIP~\cite{laptev2005space} and GMM-HMM~\cite{tang2015real} are traditional methods based on hand-crafted features. 
GLE-Net \cite{hu2020global} still achieves the best performance. 
Notably, our method achieves the best performance, reaching $97.6\%$ top-1 accuracy.

\noindent \textbf{Evaluation on MSASL.}
MSASL brings new challenges due to unconstrained recording settings.
As shown in Table~\ref{msasl}, compared with the RGB baseline~\cite{joze2018ms}, ST-GCN~\cite{yan2018spatial} shows inferior performance. 
It may be caused by the failure of pose detection on sign videos, which contains the partially occluded upper body, motion blur and noisy backgrounds. 
Albanie \emph{et al.}~\cite{albanie2020bsl} and Li~\emph{et al.} \cite{li2020transfer} both use more external RGB sign data to boost the performance on MSASL or its subsets. 
It is worth noting that our method achieves noticeable performance improvement when compared with both pose-based and RGB-based methods.

\noindent \textbf{Evaluation on WLASL.}
Compared with MSASL, WLASL contains fewer samples and double vocabulary size. 
It can be observed that Ours~(H + P), which only utilizes pose as the input modality, even outperforms the most challenging RGB-based method~\cite{albanie2020bsl}.
Besides, Ours~(H + R) further outperforms the best competitor by $7.87\%$ per-instance top-1 accuracy improvement on WLASL2000.
With incorporated hand prior and self-supervised pre-training, our method is more effective under the benchmark with limited samples.

\section{Conclusion}
In this paper, we introduce the \emph{first} self-supervised pre-trainable SLR framework with model-aware hand prior incorporated, namely SignBERT.
We involve both hands and view hand pose as a visual token.
The visual token is embedded with gesture state, temporal and hand chirality information before feeding into the framework.
We first perform self-supervised pre-training on a large volume of hand poses by masking and reconstructing the hand tokens.
During pre-training, our framework consists of the Transformer encoder and hand-model-aware decoder.
Jointly with incorporated hand prior by the decoder, we elaborately design several masking strategies for better capturing hierarchical contextual information.
Then our pre-trained framework is fine-tuned to perform recognition.
We perform extensive experiments on four popular benchmark datasets.
Experiment results demonstrate the effectiveness of our method, achieving new state-of-the-art performance on all benchmarks with a notable margin.

\footnotesize {\flushleft \bf Acknowledgements}.
This work was supported in part by the National Natural Science Foundation of China under Contract U20A20183, 61632019, and 62021001, and in part by the Youth Innovation Promotion Association CAS under Grant 2018497. It was also supported by the GPU cluster built by MCC Lab of Information Science and Technology Institution, USTC.

{\small
\bibliographystyle{ieee_fullname}
\bibliography{egbib}

\begin{thebibliography}{10}\itemsep=-1pt

\bibitem{albanie2020bsl}
Samuel Albanie, G{\"u}l Varol, Liliane Momeni, Triantafyllos Afouras, Joon~Son
  Chung, Neil Fox, and Andrew Zisserman.
\newblock {BSL}-1k: Scaling up co-articulated sign language recognition using
  mouthing cues.
\newblock In {\em ECCV}, pages 35--53, 2020.

\bibitem{ballan2012motion}
Luca Ballan, Aparna Taneja, J{\"u}rgen Gall, Luc Van~Gool, and Marc Pollefeys.
\newblock Motion capture of hands in action using discriminative salient
  points.
\newblock In {\em ECCV}, pages 640--653, 2012.

\bibitem{boukhayma20193d}
Adnane Boukhayma, Rodrigo~de Bem, and Philip~HS Torr.
\newblock {3D} hand shape and pose from images in the wild.
\newblock In {\em CVPR}, pages 10843--10852, 2019.

\bibitem{cai2019exploiting}
Yujun Cai, Liuhao Ge, Jun Liu, Jianfei Cai, Tat-Jen Cham, Junsong Yuan, and
  Nadia~Magnenat Thalmann.
\newblock Exploiting spatial-temporal relationships for {3D} pose estimation
  via graph convolutional networks.
\newblock In {\em ICCV}, pages 2272--2281, 2019.

\bibitem{camgoz2017subunets}
Necati~Cihan Camgoz, Simon Hadfield, Oscar Koller, and Richard Bowden.
\newblock {SubUNets}: End-to-end hand shape and continuous sign language
  recognition.
\newblock In {\em ICCV}, pages 3075--3084, 2017.

\bibitem{camgoz2020sign}
Necati~Cihan Camgoz, Oscar Koller, Simon Hadfield, and Richard Bowden.
\newblock Sign language {Transformers}: Joint end-to-end sign language
  recognition and translation.
\newblock In {\em CVPR}, pages 10023--10033, 2020.

\bibitem{cao2018skeleton}
Congqi Cao, Cuiling Lan, Yifan Zhang, Wenjun Zeng, Hanqing Lu, and Yanning
  Zhang.
\newblock Skeleton-based action recognition with gated convolutional neural
  networks.
\newblock {\em TCSVT}, 29(11):3247--3257, 2018.

\bibitem{carreira2017quo}
Joao Carreira and Andrew Zisserman.
\newblock Quo vadis, action recognition? {A} new model and the {Kinetics}
  dataset.
\newblock In {\em CVPR}, pages 6299--6308, 2017.

\bibitem{chen2020generative}
Mark Chen, Alec Radford, Rewon Child, Jeffrey Wu, Heewoo Jun, David Luan, and
  Ilya Sutskever.
\newblock Generative pretraining from pixels.
\newblock In {\em ICML}, pages 1691--1703, 2020.

\bibitem{cheng2020fully}
Ka~Leong Cheng, Zhaoyang Yang, Qifeng Chen, and Yu-Wing Tai.
\newblock Fully convolutional networks for continuous sign language
  recognition.
\newblock In {\em ECCV}, pages 697--714, 2020.

\bibitem{mmpose2020}
MMPose Contributors.
\newblock {OpenMMLab} pose estimation toolbox and benchmark.
\newblock \url{https://github.com/open-mmlab/mmpose}, 2020.

\bibitem{cui2019deep}
Runpeng Cui, Hu Liu, and Changshui Zhang.
\newblock A deep neural framework for continuous sign language recognition by
  iterative training.
\newblock {\em TMM}, 21(7):1880--1891, 2019.

\bibitem{deng2009imagenet}
Jia Deng, Wei Dong, Richard Socher, Li-Jia Li, Kai Li, and Li Fei-Fei.
\newblock {ImageNet}: A large-scale hierarchical image database.
\newblock In {\em CVPR}, pages 248--255, 2009.

\bibitem{devlin2018bert}
Jacob Devlin, Ming-Wei Chang, Kenton Lee, and Kristina Toutanova.
\newblock {BERT}: Pre-training of deep bidirectional transformers for language
  understanding.
\newblock In {\em NAACL}, pages 4171--4186, 2018.

\bibitem{du2015hierarchical}
Yong Du, Wei Wang, and Liang Wang.
\newblock Hierarchical recurrent neural network for skeleton based action
  recognition.
\newblock In {\em CVPR}, pages 1110--1118, 2015.

\bibitem{duan2020omni}
Haodong Duan, Yue Zhao, Yuanjun Xiong, Wentao Liu, and Dahua Lin.
\newblock Omni-sourced webly-supervised learning for video recognition.
\newblock In {\em ECCV}, pages 670--688, 2020.

\bibitem{feichtenhofer2019slowfast}
Christoph Feichtenhofer, Haoqi Fan, Jitendra Malik, and Kaiming He.
\newblock Slowfast networks for video recognition.
\newblock In {\em ICCV}, pages 6202--6211, 2019.

\bibitem{habermann2020deepcap}
Marc Habermann, Weipeng Xu, Michael Zollhofer, Gerard Pons-Moll, and Christian
  Theobalt.
\newblock {DeepCap}: Monocular human performance capture using weak
  supervision.
\newblock In {\em CVPR}, pages 5052--5063, 2020.

\bibitem{hu2021model}
Hezhen Hu, Weilun Wang, Wengang Zhou, Weichao Zhao, and Houqiang Li.
\newblock Model-aware gesture-to-gesture translation.
\newblock In {\em CVPR}, pages 16428--16437, 2021.

\bibitem{hu2021hand}
Hezhen Hu, Wengang Zhou, and Houqiang Li.
\newblock Hand-model-aware sign language recognition.
\newblock In {\em AAAI}, pages 1558--1566, 2021.

\bibitem{hu2020global}
Hezhen Hu, Wengang Zhou, Junfu Pu, and Houqiang Li.
\newblock Global-local enhancement network for {NMFs-aware} sign language
  recognition.
\newblock {\em TOMM}, 17(3):1--18, 2021.

\bibitem{huang2018attention}
Jie Huang, Wengang Zhou, Houqiang Li, and Weiping Li.
\newblock Attention based {3D-CNNs} for large-vocabulary sign language
  recognition.
\newblock {\em TCSVT}, 29(9):2822--2832, 2019.

\bibitem{huang2018video}
Jie Huang, Wengang Zhou, Qilin Zhang, Houqiang Li, and Weiping Li.
\newblock Video-based sign language recognition without temporal segmentation.
\newblock In {\em AAAI}, pages 2257--2264, 2018.

\bibitem{joze2018ms}
Hamid Reza~Vaezi Joze and Oscar Koller.
\newblock {MS-ASL}: A large-scale data set and benchmark for understanding
  american sign language.
\newblock {\em BMVC}, pages 1--16, 2019.

\bibitem{kavan2005spherical}
Ladislav Kavan and Ji{\v{r}}{\'\i} {\v{Z}}{\'a}ra.
\newblock Spherical blend skinning: a real-time deformation of articulated
  models.
\newblock In {\em ACM I3D}, pages 9--16, 2005.

\bibitem{kiros2015skip}
Ryan Kiros, Yukun Zhu, Ruslan Salakhutdinov, Richard~S Zemel, Antonio Torralba,
  Raquel Urtasun, and Sanja Fidler.
\newblock Skip-thought vectors.
\newblock In {\em NeurIPS}, pages 3294--3302, 2015.

\bibitem{koller2020quantitative}
Oscar Koller.
\newblock Quantitative survey of the state of the art in sign language
  recognition.
\newblock {\em arXiv}, pages 1--40, 2020.

\bibitem{koller2019weakly}
Oscar Koller, Cihan Camgoz, Hermann Ney, and Richard Bowden.
\newblock Weakly supervised learning with multi-stream {CNN-LSTM-HMMs} to
  discover sequential parallelism in sign language videos.
\newblock {\em TPAMI}, 42(9):2306--2320, 2020.

\bibitem{koller2018deep}
Oscar Koller, Sepehr Zargaran, Hermann Ney, and Richard Bowden.
\newblock Deep sign: Enabling robust statistical continuous sign language
  recognition via hybrid cnn-hmms.
\newblock {\em IJCV}, 126(12):1311--1325, 2018.

\bibitem{laptev2005space}
Ivan Laptev.
\newblock On space-time interest points.
\newblock {\em IJCV}, 64(2-3):107--123, 2005.

\bibitem{lewis2000pose}
John~P Lewis, Matt Cordner, and Nickson Fong.
\newblock Pose space deformation: a unified approach to shape interpolation and
  skeleton-driven deformation.
\newblock In {\em SIGGRAPH}, pages 165--172, 2000.

\bibitem{li2018co}
Chao Li, Qiaoyong Zhong, Di Xie, and Shiliang Pu.
\newblock Co-occurrence feature learning from skeleton data for action
  recognition and detection with hierarchical aggregation.
\newblock pages 786--792, 2018.

\bibitem{li2020transfer}
Dongxu Li, Cristian Rodriguez, Xin Yu, and Hongdong Li.
\newblock Transferring cross-domain knowledge for video sign language
  recognition.
\newblock In {\em CVPR}, pages 6205--6214, 2020.

\bibitem{li2020word}
Dongxu Li, Cristian Rodriguez, Xin Yu, and Hongdong Li.
\newblock Word-level deep sign language recognition from video: A new
  large-scale dataset and methods comparison.
\newblock In {\em WACV}, pages 1459--1469, 2020.

\bibitem{li2020unicoder}
Gen Li, Nan Duan, Yuejian Fang, Ming Gong, and Daxin Jiang.
\newblock Unicoder-{VL}: A universal encoder for vision and language by
  cross-modal pre-training.
\newblock In {\em AAAI}, pages 11336--11344, 2020.

\bibitem{lin2019tsm}
Ji Lin, Chuang Gan, and Song Han.
\newblock {TSM}: Temporal shift module for efficient video understanding.
\newblock In {\em ICCV}, pages 7083--7093, 2019.

\bibitem{min2020efficient}
Yuecong Min, Yanxiao Zhang, Xiujuan Chai, and Xilin Chen.
\newblock An efficient pointlstm for point clouds based gesture recognition.
\newblock In {\em CVPR}, pages 5761--5770, 2020.

\bibitem{oikonomidis2014evolutionary}
Iason Oikonomidis, Manolis~IA Lourakis, and Antonis~A Argyros.
\newblock Evolutionary quasi-random search for hand articulations tracking.
\newblock In {\em CVPR}, pages 3422--3429, 2014.

\bibitem{paszke2019pytorch}
Adam Paszke, Sam Gross, Francisco Massa, Adam Lerer, James Bradbury, Gregory
  Chanan, Trevor Killeen, Zeming Lin, Natalia Gimelshein, Luca Antiga, et~al.
\newblock {PyTorch}: An imperative style, high-performance deep learning
  library.
\newblock In {\em NeurIPS}, pages 1--12, 2019.

\bibitem{pennington2014glove}
Jeffrey Pennington, Richard Socher, and Christopher~D Manning.
\newblock Glove: Global vectors for word representation.
\newblock In {\em EMNLP}, pages 1532--1543, 2014.

\bibitem{qian2014realtime}
Chen Qian, Xiao Sun, Yichen Wei, Xiaoou Tang, and Jian Sun.
\newblock Realtime and robust hand tracking from depth.
\newblock In {\em CVPR}, pages 1106--1113, 2014.

\bibitem{qiu2017learning}
Zhaofan Qiu, Ting Yao, and Tao Mei.
\newblock Learning spatio-temporal representation with pseudo-{3D} residual
  networks.
\newblock In {\em ICCV}, pages 5533--5541, 2017.

\bibitem{radford2018improving}
Alec Radford, Karthik Narasimhan, Tim Salimans, and Ilya Sutskever.
\newblock Improving language understanding by generative pre-training.
\newblock {\em arxiv}, pages 1--12, 2018.

\bibitem{romero2017embodied}
Javier Romero, Dimitrios Tzionas, and Michael~J Black.
\newblock Embodied hands: Modeling and capturing hands and bodies together.
\newblock {\em ToG}, 36(6):1--17, 2017.

\bibitem{song2017end}
Sijie Song, Cuiling Lan, Junliang Xing, Wenjun Zeng, and Jiaying Liu.
\newblock An end-to-end spatio-temporal attention model for human action
  recognition from skeleton data.
\newblock In {\em AAAI}, pages 4263--4270, 2017.

\bibitem{sridhar2013interactive}
Srinath Sridhar, Antti Oulasvirta, and Christian Theobalt.
\newblock Interactive markerless articulated hand motion tracking using {RGB}
  and depth data.
\newblock In {\em ICCV}, pages 2456--2463, 2013.

\bibitem{Su2020VLBERT}
Weijie Su, Xizhou Zhu, Yue Cao, Bin Li, Lewei Lu, Furu Wei, and Jifeng Dai.
\newblock {VL-BERT}: Pre-training of generic visual-linguistic representations.
\newblock In {\em ICLR}, pages 1--16, 2020.

\bibitem{sun2019videobert}
Chen Sun, Austin Myers, Carl Vondrick, Kevin Murphy, and Cordelia Schmid.
\newblock {VideoBERT}: A joint model for video and language representation
  learning.
\newblock In {\em ICCV}, pages 7464--7473, 2019.

\bibitem{tang2015real}
Ao Tang, Ke Lu, Yufei Wang, Jie Huang, and Houqiang Li.
\newblock A real-time hand posture recognition system using deep neural
  networks.
\newblock {\em ACM TIST}, 6(2):1--23, 2015.

\bibitem{tkach2016sphere}
Anastasia Tkach, Mark Pauly, and Andrea Tagliasacchi.
\newblock Sphere-meshes for real-time hand modeling and tracking.
\newblock {\em ToG}, 35(6):1--11, 2016.

\bibitem{tunga2020pose}
Anirudh Tunga, Sai~Vidyaranya Nuthalapati, and Juan Wachs.
\newblock Pose-based sign language recognition using {GCN} and {BERT}.
\newblock In {\em WACV Workshop}, pages 31--40, 2020.

\bibitem{tzionas2016capturing}
Dimitrios Tzionas, Luca Ballan, Abhilash Srikantha, Pablo Aponte, Marc
  Pollefeys, and Juergen Gall.
\newblock Capturing hands in action using discriminative salient points and
  physics simulation.
\newblock {\em IJCV}, 118(2):172--193, 2016.

\bibitem{vaswani2017attention}
Ashish Vaswani, Noam Shazeer, Niki Parmar, Jakob Uszkoreit, Llion Jones,
  Aidan~N Gomez, Lukasz Kaiser, and Illia Polosukhin.
\newblock Attention is all you need.
\newblock In {\em NeurIPS}, pages 5999--6009, 2017.

\bibitem{wu2016google}
Yonghui Wu, Mike Schuster, Zhifeng Chen, Quoc~V Le, Mohammad Norouzi, Wolfgang
  Macherey, Maxim Krikun, Yuan Cao, Qin Gao, Klaus Macherey, et~al.
\newblock Google's neural machine translation system: Bridging the gap between
  human and machine translation.
\newblock {\em arXiv}, pages 1--23, 2016.

\bibitem{yan2018spatial}
Sijie Yan, Yuanjun Xiong, and Dahua Lin.
\newblock Spatial temporal graph convolutional networks for skeleton-based
  action recognition.
\newblock In {\em AAAI}, pages 7444--7452, 2018.

\bibitem{yang2019xlnet}
Zhilin Yang, Zihang Dai, Yiming Yang, Jaime Carbonell, Ruslan Salakhutdinov,
  and Quoc~V Le.
\newblock {XLNet}: Generalized autoregressive pretraining for language
  understanding.
\newblock In {\em NeurIPS}, pages 1--18, 2019.

\bibitem{yuan20172017}
Shanxin Yuan, Qi Ye, Guillermo Garcia-Hernando, and Tae-Kyun Kim.
\newblock The 2017 hands in the million challenge on {3D} hand pose estimation.
\newblock {\em arXiv}, pages 1--7, 2017.

\bibitem{zhang2017hand}
Jiawei Zhang, Jianbo Jiao, Mingliang Chen, Liangqiong Qu, Xiaobin Xu, and
  Qingxiong Yang.
\newblock A hand pose tracking benchmark from stereo matching.
\newblock In {\em ICIP}, pages 982--986, 2017.

\bibitem{zhou2021improving}
Hao Zhou, Wengang Zhou, Weizhen Qi, Junfu Pu, and Houqiang Li.
\newblock Improving sign language translation with monolingual data by sign
  back-translation.
\newblock In {\em CVPR}, pages 1316--1325, 2021.

\bibitem{Zhu_2020_CVPR}
Linchao Zhu and Yi Yang.
\newblock {ActBERT}: Learning global-local video-text representations.
\newblock In {\em CVPR}, pages 8746--8755, 2020.

\bibitem{zimmermann2017learning}
Christian Zimmermann and Thomas Brox.
\newblock Learning to estimate {3D} hand pose from single {RGB} images.
\newblock In {\em ICCV}, pages 4903--4911, 2017.

\end{thebibliography}
}

\end{document}